# A Hybrid Approach of Transfer Learning and Physics-Informed Modeling: Improving Dissolved Oxygen Concentration Prediction in an Industrial Wastewater Treatment Plant

Ece Serenat Koksal[a], Erdal Aydin[a,b,*]

[a]*Department of Chemical and Biological Engineering, Koç University, Istanbul 34450, Turkey*

[b]*Koç University TUPRAS Energy Center (KUTEM), Koç University, Istanbul, 34450, Turkey*

**ABSTRACT**

Constructing first principles models is a challenging task for nonlinear and complex systems such as a wastewater treatment unit. In recent years, data-driven models are widely used to overcome the complexity. However, they often suffer from issues such as missing, low quality or noisy data. Transfer learning is a solution for this issue where knowledge from another task is transferred to target one to increase the prediction performance. In this work, the objective is increasing the prediction performance of an industrial wastewater treatment plant by transferring the knowledge of (i) an open-source simulation model that captures the underlying physics of the process, albeit with dissimilarities to the target plant, (ii) another industrial plant characterized by noisy and limited data but located in the same refinery, and (iii) the model in (ii) and making the objective function of the training problem physics informed where the physics information derived from the open-source model in (ii). The results have shown that test and validation performance are improved up to 27% and 59%, respectively.

---

[*] Corresponding Author.
E-mail adress: eaydin@ku.edu.tr (E. Aydin).

## 1. Introduction

Artificial neural networks (ANN) are unique type of mathematical representations of input and output relations. In addition to an input and an output layer, there are hidden layers, user defined activation functions representing the neurons with certain amount. The recurrent neural networks (RNN) are dynamic structures of ANNs where the past information is also considered. RNNs are suited especially for dynamic process with time-series data. However, they often suffer from an issue called "vanishing gradient problem" meaning the gradients become smaller as they are backpropagated from the output layer to the previous layers. As a respond to this problem, long short-term memory (LSTM) blocks are widely used structures where the information flow is controlled by gates [1].

Data driven models are widely used since they are easy to construct compared to first-principles models. Considering the complexity of the first principles models, data driven models are good alternatives to model processes when data is present. However, these models are black box in nature often suffer from issues such as overfitting, underfitting, lack of data or noisy data [2]. As a solution, physics-informed neural networks are suggested in many works where physics-based knowledge is integrated to the data driven models to response to the aforementioned issues [3]–[11].

Transfer learning is transferring the knowledge of a related task with abundant data to the target task where the data is scarce to increase the prediction performance. Based on the solution categorization, transfer learning can be classified into four: instance, feature, parameter, relational based approaches. The parameter-based transfer learning focuses on transferring the knowledge at the model and the parameter level which is basically transferring the weight and bias information at the layers of the machine learning model [12], [13]. Transfer learning is used for many areas such as image classification, drug discovery, fault diagnosis, radioactive particle tracking and skin cancer classification [14]–[18]. Yet, its implementation to regression problems is very limited in the literature, especially for chemical and environmental engineering tasks.

The similarity of the features between the related task and the main task can be quite important in transfer learning. Li & Rangarajan used interpretable linear models for molecular property and demonstrated the requirement of overlapping features with the transferred information and the model. While transfer learning can be detrimental when percentage of overlapping features is low, the accuracy and generalizability is higher when knowledge is transferred from a task with higher percentage of overlapping features [19].

Transfer learning has been applied to increase prediction capability and overcome the issues of lack of useful data in chemical engineering or related applications such as gas adsorption in metal-organic frameworks and industrial distillation columns in the literature [20], [21]. Chuang et al. created a base model which mimics Computational Fluid Dynamics (CFD) simulator to determine the feasible operating region. They collected data from the CFD and creates a final model by using transfer learning based on Bayesian migration technique. The contribution of transfer learning provided constructing a final model with small dataset size [22]. Rogers et al. demonstrated two case studies where different transfer learning strategies sustain more accuracy and reliability than kinetic and fully data driven models of biochemical processes [23]. Wu & Zhao proposed a fault detection and diagnosis method based on transfer learning for multimode chemical processes and used Tennessee Eastman process as case study. Information transferred from source mode eliminated the issue of lack of data and it is useful even when data is unlabeled [24]. Bi et al. integrated transfer learning modeling and intelligent heuristic optimization for a steam cracking process to decrease the time consumption and data size requirements [25].

Transfer learning is also applied to wastewater treatment plants in which biological treatment is achieved with activated sludge. Pisa et al. trained and tested an LSTM based Proportional-Integral (PI) controller to capture the quickest dynamics. Then, they transferred the LSTM based controller's knowledge to a different control loop. They obtained a final controller with an improved performance and reduced design complexity [26]. Huang et al. considered the knowledge of ErWu wastewater treatment plant as source domain and transfer it to increase the prediction capacity of TuoDong wastewater treatment plant based on chemical oxygen demand (COD) by using two-stage TrAdaBoost.R2 algorithm [27].

Physics informed neural networks and transfer learning approaches are widely used to increase the prediction performance and obtain more reliable models. Recently, their combination is also applied by some research which enables to improve the data-driven models further by using the advantage of both approaches. Lansford et al. integrated transfer learning with physics informed directed message passing neural network to predict out-of-sample vapor pressure [28]. Guc & Chen utilized physics informed transfer learning through GoogLeNet structure for fault cause assignment [29]. Schröder et al. improved wind farm monitoring by transferring simulation data and adding a physical constraint to the artificial neural network model [30]. Goswami et al. utilized transfer learning enhanced physics informed neural networks to solve brittle fracture problems [31]. Xu et al. proposed physics informed transfer learning to solve inverse problems in linear elasticity and hyper elasticity. Then, in the online stage, utilized transfer learning for fine-tuning to deal with the noisy data [32].

In the landscape of wastewater treatment modeling, prior studies have predominantly focused on classical machine learning or first-principles models. However, classical machine learning approaches may encounter limitations in effectively modeling real industrial data, and the construction of first-principles models proves to be a challenging and time-intensive endeavor. What sets our work apart is the integration of transfer learning, a technique not extensively explored in the context of industrial wastewater treatment. By addressing this gap, our research pioneers a holistic approach that not only increases the prediction performance but also serves as a steppingstone for further advancements in the application of transfer learning to complex industrial processes. In this work, prediction performance of an industrial wastewater treatment unit is increased by transferring the knowledge of (i) an open-source simulation model, (ii) another industrial wastewater treatment plant in the same refinery and (iii) combining the physics knowledge based on (i) and integrating it to (ii). The key contribution of the proposed work lies on (i) combining transfer learning with LSTM models for regression problems related to process engineering, (ii)increasing the prediction performance of a real plant by the knowledge of an open-source model, (iii) increasing the prediction performance of an actual plant by another real plant although the source model is trained with noisy data with low set size, (iv) applying physics informed transfer learning with recurrent neural network structure and measuring its performance on an actual plant.

The paper is organized as follows: A brief introduction for neural networks and transfer learning is expressed in Section 1. Section 2 explains the open-source model and the industrial wastewater treatment plant in interest. The methodology related to the machine learning methods and transfer learning is explained in Section 3. Section 4 presents three case studies illustrating the application of transfer learning to increase the prediction performance of the target model. Finally, Section 5 offers concluding remarks to summarize the key findings and insights presented in this paper.

## 2. Process Model: Wastewater Treatment Unit

Wastewater treatment is the process designed to clean the wastewater before it is discharged to the environment. Wastewater treatment is crucial for removing contaminants, pollutants and impurities since the quality of the treated water affects the environment and public health. In wastewater treatment process, the wastewater passes through physical, chemical and biological treatment before being discharged to the environment.

Activated sludge is the biological treatment process where the bacteria content breaks down the organic matter in the wastewater. Modeling wastewater treatment systems can be quite challenging due to nonlinearity as a result of parallel reactions. In 1983, International Association of Water Pollution Research and Control (IAWPRC) formed a group to construct a base model for single sludge systems to facilitate the design and operation of wastewater treatment units. To respond this need, Henze et al. designed a general model for single-sludge wastewater treatment systems named as Activated Sludge Model 1 (ASM1) representing the model development for carbon oxidation, nitrification, and denitrification [33]. Although ASM1 is subject to some limitations, it is universally accepted [34]. Based on ASM1, Alex et al. printed Benchmark Simulation Model 1 (BSM1), with biological reactor and secondary clarifier sections of a wastewater treatment unit where the biological reactor section consists of two anaerobic and three aerobic reactors in series form [35].

Besides the difficulties on modelling wastewater treatment systems due to the complex and nonlinear behavior, controlling dissolved oxygen concentration is quite important to ensure that the bacteria content has enough oxygen to efficiently treat the wastewater. However, feeding the unit with excess oxygen gives birth to high energy consumption. The oxygen is fed to the activated sludge by aerators and the energy consumption of aerators is between 30-75% of total energy consumption in activated sludge system [36]. As a result, predicting the dissolved oxygen concentration helps to maintain a healthy bacteria content, decrease energy consumption and operate with effective minimal cost. Lin & Luo adapted a radial basis function neural network controller to control dissolved oxygen concentration at the aerobic section of ASM1 [37]. For the same purpose, Tzoneva implemented PID controllers [38]. In addition, Holenda et al. designed a model predictive controller (MPC) for ASM1 [39]. Qambar & Al Khalidy optimized dissolved oxygen concentration at Madinat Salman wastewater treatment plant by predicting with random forest and gradient boost models resulting in lower energy consumption [40]. Asrav et al. constructed recurrent neural network models to increase the prediction performance of an industrial plant [41].

In ASM1, the change of dissolved oxygen concentration at the aerobic section of the reactors are represented as follows [33]:

$$\frac{dS_O}{dt} = -\frac{1-Y_H}{Y_H}\mu_H \left(\frac{S_S}{K_S + S_S}\right)\left(\frac{S_O}{K_{O,H} + S_O}\right) x_{BH}$$
$$- \frac{4.57 - Y_A}{Y_A}\mu_A \left(\frac{S_{NH}}{K_{NH} + S_{NH}}\right)\left(\frac{S_O}{K_{O,A} + S_O}\right) x_{BA} \quad (1)$$

where $S_O$ is the dissolved oxygen concentration at the activated sludge (g/m$^3$), $S_S$ is the readily biodegradable substrate (g COD / m$^3$), $S_{NH}$ is the NH$_4^+$+NH$_3$ nitrogen (g N / m$^3$), $x_{BH}$ is the active heterotrophic biomass (g COD/m$^3$), $x_{BA}$ is the active autotrophic biomass (g COD/m$^3$), $Y_H$ is the heterotrophic yield (g $x_{BH}$ COD formed$^{-1}$), $Y_A$ is the autotrophic yield (g $x_{BA}$ COD formed$^{-1}$), $\mu_H$ is the maximum heterotrophic growth rate (day$^{-1}$), $\mu_A$ is the maximum autotrophic growth rate (day$^{-1}$), $K_S$ is the half saturation of heterotrophic growth (g COD/m$^3$), $K_{NH}$ is the half saturation of autotrophic growth (g NH$_3$-N / m$^3$), $K_{O,H}$ is the half saturation of heterotrophic oxygen (g COD/m$^3$), $K_{O,A}$ is the half saturation of autotrophic oxygen (g O$_2$ / m$^3$). For the values of $Y_H$, $Y_A$, $\mu_H$, $\mu_A$, $K_S$, $K_{NH}$, $K_{O,H}$ and $K_{O,A}$, and for more detailed information on ASM1, readers may refer to [33].

In the wastewater treatment plant analyzed in this study, the physical treatment starts at the first unit of the process, Oil Water Separator (OWS) which is followed by chemical treatment procedure. The next unit is dissolved air floatation (DAF) where the suspended solids are removed by attaching bubbles causing them floating to the surface. The outlet stream properties such as pH, total suspended solid (TSS), chemical oxygen demand (COD), grease, ammonium nitrogen ($NH_4$-N), phenol and sulfur are measured regularly since the stream is sent to activated sludge where the biological content removes waste from the water by parallel reactions. The outlet stream of the activated sludge is separated in a specified ratio. The first separated stream is sent back to the inlet of the activated sludge system and the second one is fed to the clarifier which is the last unit in the wastewater treatment plant. Compared to ASM1, the wastewater treatment plant in interest has no anaerobic section in the activated sludge section. In addition, there are two activated sludges, and their inlet stream is separated into two at the outlet of the DAF. The representation of the process is shown in Fig. 1.

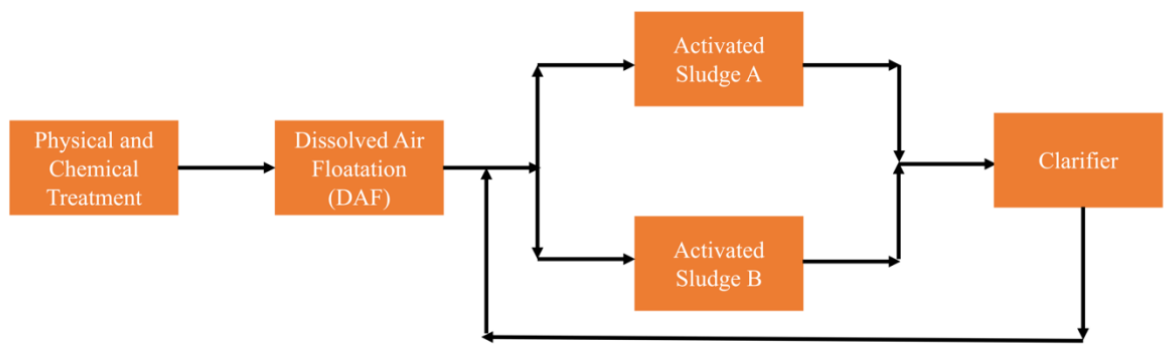

**Fig. 1.** Representation of the industrial wastewater treatment model.

## 3. Methodology

Artificial neural networks are mathematical representation of input and output relations. In an artificial neural networks structure inspired by neural connections in the brain, there are user defined number of hidden layers and neurons at each layer. A feed-forward and fully connected artificial neural network can be represented as:

$$\hat{y} = f_1(w_1 f_2(w_2 x + b_2) + b_1) \qquad (2)$$

where $x$ is input, $\hat{y}$ is the predicted output, $f_1$ and $f_2$ are activation functions, $w_1$ and $w_2$ are weight terms, $b_1$ and $b_2$ are bias terms for output and hidden layers, respectively. For a recurrent neural network, the mathematical expression of the predicted output is as follows:

$$h_t = f_2(w_2 u_t + w_2 h_{t-1} + b_2) \qquad (3a)$$
$$\hat{y}_t = f_1(w_1 f_2(w_2 x_t + w_2 h_{t-1} + b_2) + b_1) \qquad (3b)$$

where $t$ is the time step and $h_t$ and $\hat{y}_t$ are hidden state and current state output, respectively. The structure of a simple RNN is represented in Fig. 2 [42].

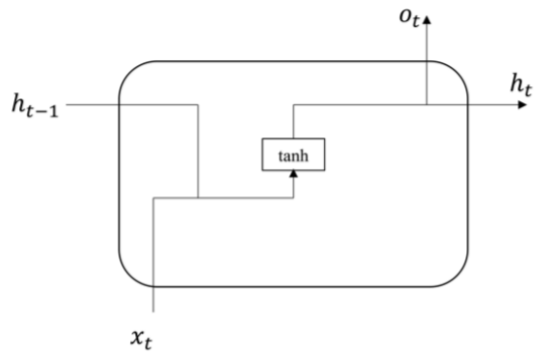

**Fig. 2.** Simple RNN structure.

Recurrent neural networks often suffer from "vanishing gradient problem" in which gradients become extremely small as they are backpropagated from output layer to previous layers [43]. As a solution to this issue, LSTM blocks are widely used where the flow of information and gradients are controlled by three gates of of the each cell, named input, forget and output gates. LSTM equations are listed as: [1]

$$i_t = \sigma(w_i x_t + u_i h_{t-1} + b_i) \tag{4a}$$

$$f_t = \sigma(w_f x_t + u_f h_{t-1} + b_f) \tag{4b}$$

$$o_t = \sigma(w_o x_t + u_o h_{t-1} + b_o) \tag{4c}$$

$$\tilde{c}_t = \tanh(w_c x_t + u_c h_{t-1} + b_c) \tag{4d}$$

$$c_t = f_t \odot c_{t-1} + i_t \odot \tilde{c}_t \tag{4e}$$

$$h_t = o_t \odot tanh(c_t) \tag{4f}$$

where $\sigma$ is the sigmoid function, $f_t$ is the forget gate activation vector, $i_t$ is the input gate activation factor, $o_t$ is the output gate activation factor, $c_t$ is the cell state vector, $\tilde{c}_t$ is the cell input activation vector, $\odot$ is the Hadamart product, $w_i, w_f, w_c, u_i, u_f, u_c, b_i, b_f, b_c$ are weight and bias terms.

In addition, gated recurrent units (GRU) are another version of RNNs consisting of two gates called reset and update without having a cell. The equations for a GRU structure are given below [44]:

$$z_t = \sigma(w_z x_t + u_z h_{t-1} + b_z) \tag{5a}$$

$$r_t = \sigma(w_r x_t + u_r h_{t-1} + b_r) \tag{5b}$$

$$\tilde{h}_t = tanh(w_h x_t + u_h(r_t \odot h_{t-1}) + b_h) \tag{5c}$$

$$h_t = (1 - z_t) \odot h_{t-1} + z_t \odot \tilde{h}_t \tag{5d}$$

where $h_t$ is the output vector, $\tilde{h}_t$ is the candidate activation vector, $z_t$ is the update gate vector, and $r_t$ is the reset gate vector. GRU has a simpler structure compared to LSTM. As a result, GRU has less parameters and computationally more advantageous. However, GRU structure may not be as efficient as LSTM to preserve the long-term dependencies. Therefore, it is more suitable for the smaller datasets [45]. The structure of the LSTM and GRU are represented in Fig. 3 [42].

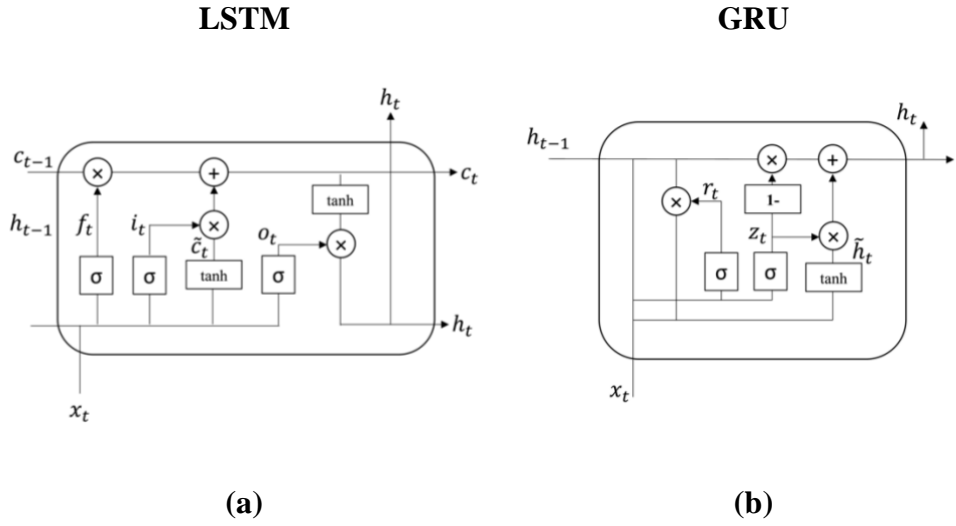

(a)          (b)

**Fig. 3.** LSTM and GRU structure.

One way to evaluate the training and test performance of the models is mean squared error (MSE) which is calculated as follows:

$$\frac{1}{N}\sum_{i=1}^{N}(\hat{y}_i - y_i)^2 \qquad (6)$$

Mean absolute error (MAE) is another metric to evaluate the performance of the models. MAE is calculated as follows:

$$MAE = \frac{1}{N}\sum_{i=1}^{N}|\hat{y}_i - y_i| \tag{7}$$

Training is an optimization problem in which MSE is minimized. During training an LSTM network, the objective function of the training problem becomes:

$$min_{w_i,w_f,w_c,u_i,u_f,u_c,b_i,b_f,b_c} \frac{1}{N}\sum_{i=1}^{N}(\hat{y}_i - y_i)^2 \tag{8}$$

Physics informed neural networks are widely used to increase the prediction performance and to handle the issues related to the black-box nature of neural network training. One way to make a neural network is embedding a physics term to the objective function so that the neural network minimizes the error and obeys the physics rule simultaneously. In case of static neural networks, the change in the objective function is straightforward:

$$min_{w_1,w_2,b_1,b_1} \frac{1}{N}\sum_{i=1}^{N}(\hat{y}_i - y_i)^2 + (\alpha \boldsymbol{p}) \tag{9}$$

where $\boldsymbol{p}$ is the physics term and $\alpha$ is the weight for the physics term.

Recurrent neural networks are dynamic structures governed by differential equations. In order to embed a physics term to the objective function, firstly the discretized form of the differential equation according to the Euler backward method is used:

$$y_n = y_{n-1} + (t_n - t_{n-1})f(x_n, y_n, t_n) \tag{10}$$

4where t is the time and n is the time step. Rewriting the expression so that the left and the right-hand sides represent the actual and the predicted values, respectively:

$$y_n - y_{n-1} = (t_n - t_{n-1})f(x_n, y_n, t_n) \qquad (11)$$

As a result, the physics term can be written regarding that the square of the difference between the left and the right hand-sides of the Eq.11 should be minimized. Then, physics term for the recurrent neural network structures, represented as $\boldsymbol{p_r}$ becomes:

$$\boldsymbol{p_r} = \frac{1}{N}\sum_{i=1}^{N}(y_i - y_{i-1} - (t_i - t_{i-1})f(x_i, \hat{y}_i, t_i))^2 \qquad (12)$$

Embedding the physics term to the objective function and rewriting Eq. 8:

$$min_{w_i, w_f, w_c, u_i, u_f, u_c, b_i, b_f, b_c} \frac{1}{N}\sum_{i=1}^{N}(\hat{y}_i - y_i)^2 + \alpha \boldsymbol{p_r} \qquad (13)$$

Transfer learning is transferring the knowledge from a related task to the target one where data is scarce, aiming to increase the prediction performance. Based on ANN or RNN structures, transfer learning scheme can be constructed as a combination of a source model and a target model. The schematic representation of the steps of the transfer learning is given in Fig. 4.

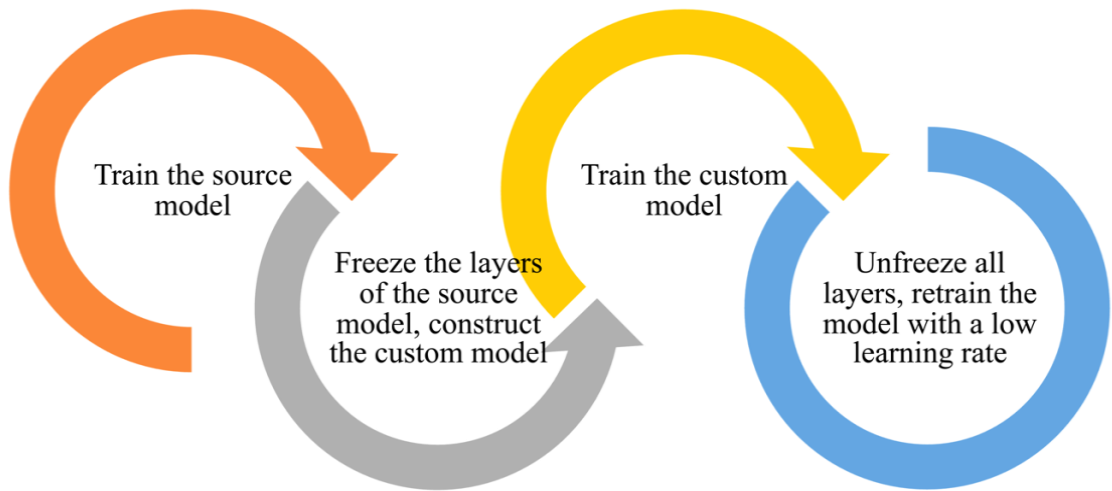

**Fig. 4.** Steps of the transfer learning.

The explanation of the steps are as follows:

*Step 1: Creating the source model*

The source model is trained aiming to achieve lowest MSE. The resulting model is called the "pretrained model".

*Step 2. Freezing the transferred layers and constructing the custom model*

The layers of the pretrained model with their weight and bias information are transferred to the target model. Additional layers are added to this model. However, the layers transferred from the pretrained model are frozen which keeps the weight and bias terms in these layers fixed. The resulting model with transferred and additional layers are called "custom model". Mathematically, this means that $w_i, w_f, w_c, u_i, u_f, u_c, b_i, b_f, b_c$ corresponding to the layers of the source model are fixed. However, these terms corresponding to the new custom model are still decision variables to be determined during the training step.

*Step 3. Training the custom model*

The custom model is trained based on the predefined hyperparameters.

*Step 4. Fine tuning*

In the last step, all layers are unfrozen. The obtained model in Step 3 is retrained with a low learning rate such as $10^{-5}$. Mathematically, this means that all $w_i, w_f, w_c, u_i, u_f, u_c, b_i, b_f, b_c$ corresponding to the source and the custom model are again decision variables but they are subject to the specified learning rate. This step enables to decreasing errors as a result of the dissimilarity between the source and the target model and adjust the weight and bias terms to obtain the final model.

The transfer learning structure is represented in Fig. 5.

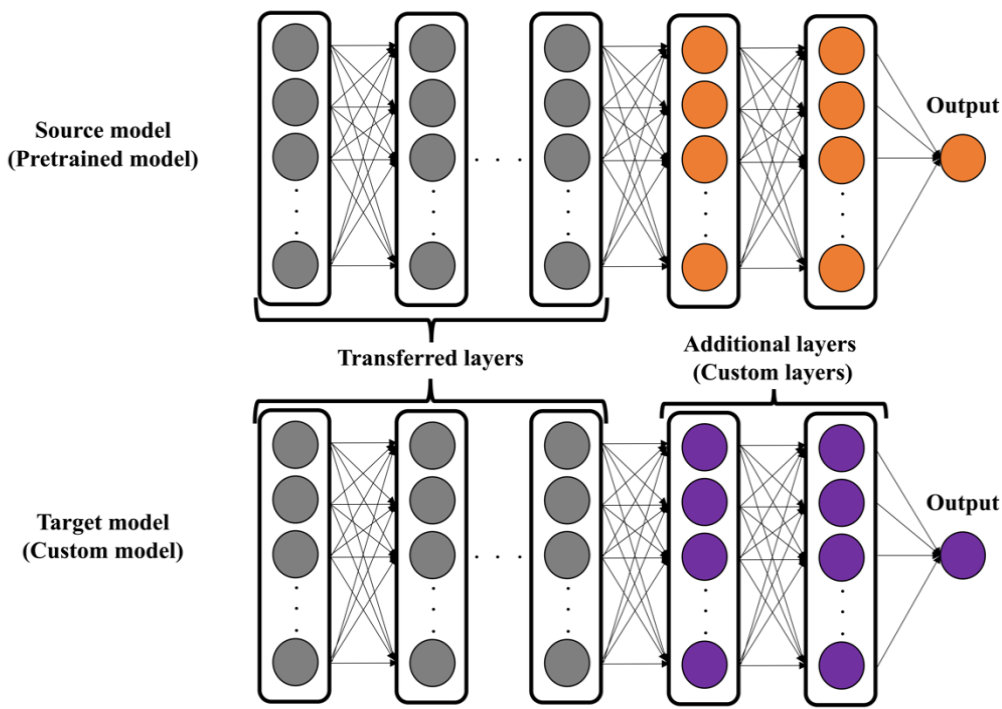

**Fig. 5.** Transfer learning structure.

Similarly, to construct a physics informed transfer learning model, only the objective function can be changed regarding the neural network structure. If static neural networks are used and the physics of the process are in algebraic relations, the physics term can be directly embedded to the objective function of the custom model. Otherwise, if recurrent neural networks are used and the physics of the process are governed by differential equations, Euler backward method provides discretized form of these relations which enables embedding the physics term to the objective function.

## 4. Results & Discussion

### 4.1. Problem Statement

Activated sludge plays a pivotal role in wastewater treatment, serving as the core of the system where biological treatment occurs, separating a significant portion of waste from water. Maintaining an optimal dissolved oxygen concentration is crucial for sustaining the aerobic bacteria responsible for biological treatment. Regular measurement of dissolved oxygen concentration is performed, and oxygen is supplied to the activated sludge system through aerators, which consume energy.

The objective of this study is to predict the dissolved oxygen concentration at activated sludge A in the target plant, as depicted in Figure 1. The features considered for prediction include DAF outlet properties (pH, TSS, COD, grease, NH4-N, phenol, sulfur), TSS at return activated sludge, and TSS at activated sludge A and B. The measurements are averaged over a 1-day interval. The training procedure is executed using the TensorFlow and Keras frameworks [46], [47], with hyperbolic tangent is chosen as the activation function for the hidden layers. The Adam optimizer algorithm is selected for training [48]. The time-step is set to 5 days in accordance with process requirements determined by the operation group. Among neural network architectures, including ANN, simple RNN, LSTM, and GRU, LSTM is chosen as the machine learning model, having demonstrated convergence to the lowest Mean Squared Error (MSE). Hyperparameters are fine-tuned through a trial-and-error process.

17At the initial stage of the research, around 700 data points were available for training and testing. Data normalization is performed between -1 and +1, and the training ratio is set at 90%. The data is sequential and not shuffled due to the dynamic nature of the process. After model construction, an offline validation procedure is executed, incorporating 200 sequential days from the end of the test dataset. Three different models with varying complexities are trained and evaluated.

The first model, termed the standard model, achieves the best test and validation performance among the trials. It is trained with 5 hidden layers, each comprising 30 neurons. The second model, more complex than the standard model, is trained with 6 hidden layers and 60 neurons. Finally, a less complex model is trained with 3 hidden layers, each comprising 20 neurons.

The training, test, and validation performance results for each model are presented in the first three rows of Tables 1 and 2. Additionally, the trends are visualized in Fig. 6. The standard model, exhibiting superior test and validation performance, serves as a benchmark for comparison with the more complex and less complex models.

**Table 1**

MSE of different models.

| MSE | Train | Test | Validation |
|---|---|---|---|
| **Standard** | 0.020 | 0.044 | 0.098 |
| **More complex** | 0.015 | 0.045 | 0.133 |
| **Less complex** | 0.026 | 0.048 | 0.110 |
| **Open-Source TL** | 0.030 | 0.037 | 0.042 |
| **Industrial TL** | 0.030 | 0.035 | 0.052 |
| **Physics Informed TL** | 0.028 | 0.032 | 0.040 |

18**Table 2**

MAE of different models.

| MAE | Train | Test | Validation |
|---|---|---|---|
| **Standard** | 0.113 | 0.155 | 0.246 |
| **More complex** | 0.098 | 0.166 | 0.255 |
| **Less complex** | 0.122 | 0.179 | 0.261 |
| **Open-Source TL** | 0.121 | 0.149 | 0.142 |
| **Industrial TL** | 0.120 | 0.141 | 0.159 |
| **Physics Informed TL** | 0.113 | 0.135 | 0.132 |

|  | **Standard and More Complex** | **Standard and Less Complex** |
|---|---|---|
| **Training** | 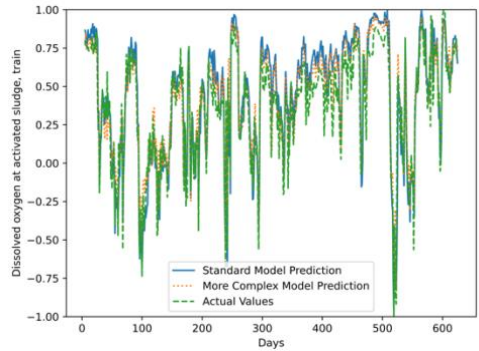 <br> (a) | 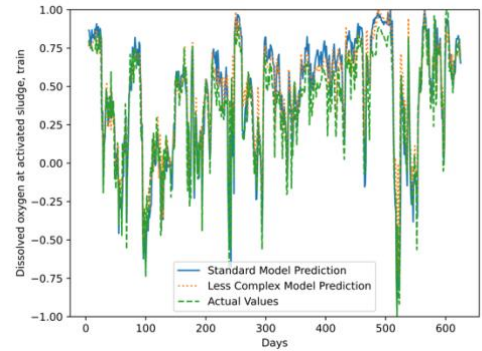 <br> (d) |
| **Test** | 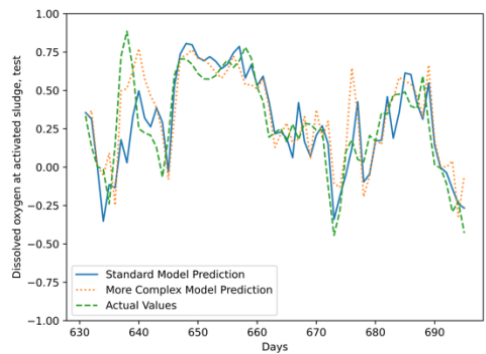 <br> (b) | 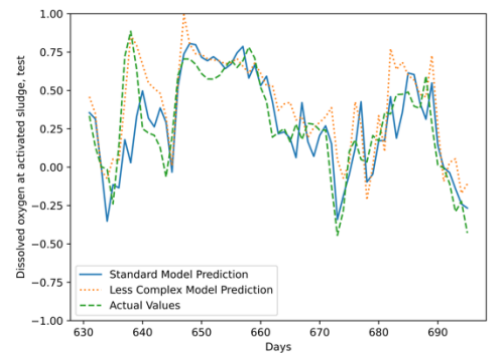 <br> (e) |

**Validation** 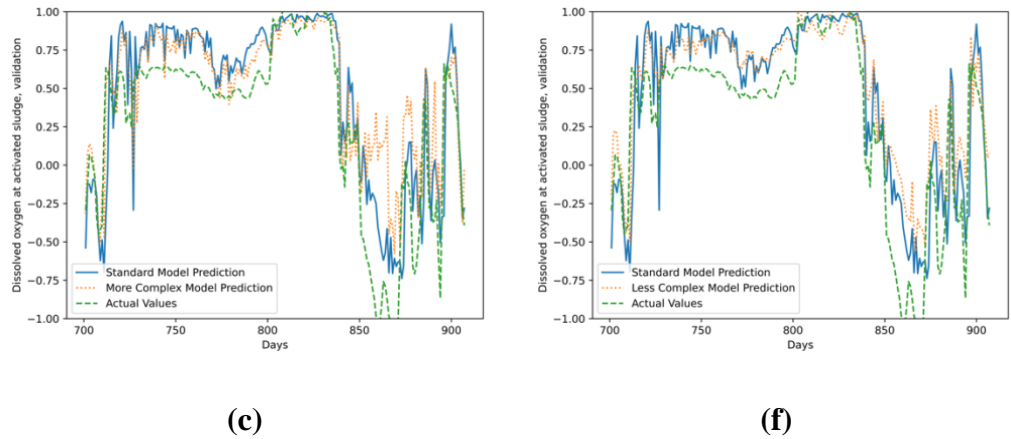

(c)                  (f)

**Fig. 6.** Trends for standard, more complex and less complex models.

The training performance of the more complex model surpasses that of the standard model, indicating a seemingly better fit to the training data. However, the test and validation MSE and MAE deteriorate as the number of layers and neurons is increased. The deviation becomes notably pronounced in the right-hand side of the validation trend, especially after the 850$^{th}$ day, where the more complex model diverges significantly from the actual trend. This divergence is accompanied by a considerable gap between the predicted and actual values, marking an overfitting of the process under consideration. The increased complexity of the model, while capturing intricacies in the training data, fails to generalize well to unseen data.

Conversely, reducing the number of layers and neurons results in an underfit, with worsened training, test, and validation performance. The deviation from actual values is once again more pronounced in the validation trend, particularly during the days 720-800 and 850-875. These outcomes underscore the delicate balance required in model complexity for accurate prediction of dissolved oxygen concentration at the activated sludge. A more nuanced approach, integrating classical machine learning with complementary methods, is necessary to achieve a more precise and robust prediction. This can involve leveraging techniques such as transfer learning or physics-informed models, as discussed in previous sections, to enhance the performance and generalization capabilities of the model.

## 4.2. Transfer Learning with an Open-Source Plant

Transfer learning indeed provides a promising avenue to address the limitations of classical machine learning and enhance prediction performance. In this study, the ASM1 model [33] is selected as the first source model to improve the prediction performance of the target model. The source model is constructed using features such as readily and slowly biodegradable substrates, particulate inert organic matter, feed flow rate, soluble and particulate biodegradable organic nitrogen, active heterotrophic biomass, and ammonium nitrogen.

For the source model, LSTM architecture is employed, featuring 5 hidden layers and 25 neurons at each layer, with the hyperbolic tangent function serving as the activation function. The dataset, comprising 673 data points, is not shuffled, as the sequential nature of the days is preserved. The resulting figures are presented in Fig. 7.a. and Fig. 7.b., with corresponding performance metrics outlined in the first column of Table 3.

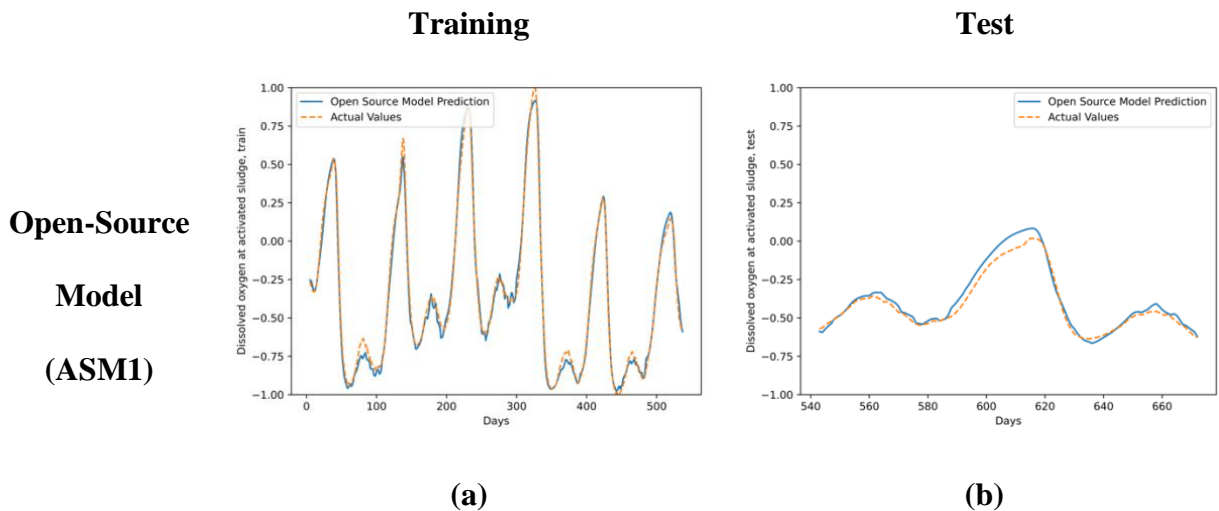

|                          | **Training** | **Test** |
|--------------------------|--------------|----------|
| **Open-Source Model (ASM1)** | (a)          | (b)      |
| **Industrial Model** | 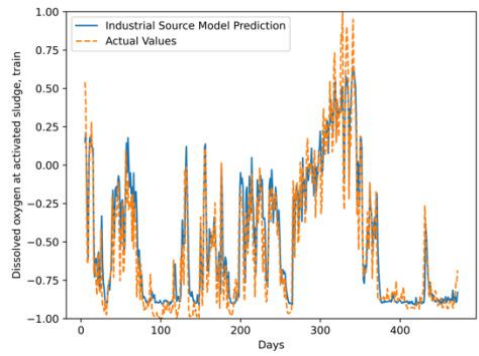 | 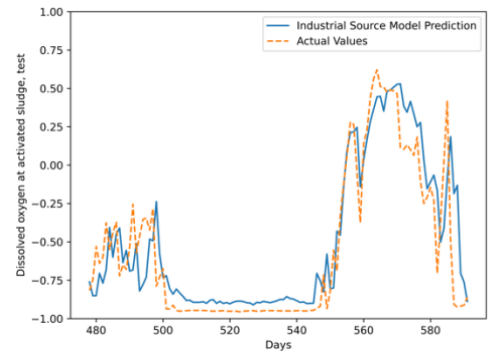 |
| | (c) | (d) |

**Fig. 7.** Trends for open-source and industrial transfer learning models.

**Table 3**

Metrics for open-source and industrial source models.

| Metric | Open-Source Model | Industrial Source Model |
|---|---|---|
| **Training MSE** | 0.0019 | 0.0262 |
| **Test MSE** | 0.0017 | 0.0485 |
| **Training MAE** | 0.0336 | 0.1167 |
| **Test MAE** | 0.0318 | 0.1550 |

In the pursuit of enhancing prediction performance, three layers of the open-source model are transferred to the target task. The custom model is designed with careful consideration to combine the source and target models, accounting for their differing numbers of features. The first layer, utilizing the identity function, includes 9 neurons, matching the number of features plus the output of the open-source model. This layer facilitates the integration of the source and target models, having potentially different type and number of features and targets.

Subsequently, the first 3 layers of the open-source model are transferred, and their weights are frozen. These frozen layers are followed by 6 additional layers, each comprising 30 neurons, where training is allowed. The output layer employs the identity function with a single output. The custom model is trained, and in the final step, fine-tuning is performed with all layers set as trainable, employing a learning rate of $10^{-5}$. The resulting model is named "Open-Source TL," with performance metrics provided in the fourth row of Tables 1 and 2. The trends for training, test, and validation are illustrated in the first column of Fig. 8.

While the training performance experiences a slight decline when layers of the open-source model are transferred, the test and validation performances see notable improvements. In terms of MSE, the test and validation performances exhibit enhancements of 15% and 36%, respectively. Correspondingly, MAE is reduced by 4% and 42% compared to the standard model for the test and validation performances. Particularly noteworthy is the strength of the usage of the open-Source TL model, which becomes more apparent in the validation set. The trends also demonstrate a reduction in the gap between actual values and the standard model prediction during the 725-800 days interval when knowledge from the open-source model is transferred. Consequently, the deviation around the $850^{th}$ day is much smaller compared to the standard model prediction.

These results might be considered as promising, since they highlight the potential of transfer learning to improve prediction performance, even when the source model has dissimilarities with the target model. Despite the challenges of industrial processes, such as intermittent inputs, noise, and data scarcity, the open-source model's knowledge, based on equation-based simulations, significantly enhances the prediction performance of the machine learning model through transfer learning. The transferred layers enable the industrial data to align with the general physics of the process, contributing to the overall improvement in prediction accuracy.

### 4.3. Transfer Learning with an Industrial Source Plant

The second case study involves transferring knowledge from another industrial plant within the same city, both equipped with wastewater treatment units. The target wastewater treatment unit is relatively new, having undergone maintenance for only a few years, while the source comprises an older plant with a dataset size of 600, smaller than the target model. Despite the smaller dataset and potential noise in measurements due to real plant conditions, the advantage lies in the similarity in structure and location between the two plants. The overarching question of this case study is whether the prediction performance of the target plant can be improved by employing a source model trained on noisy data with a smaller set size, leveraging the advantage of structural and locational similarities.

In the initial step, the source plant is trained using LSTM with an 80% training ratio, employing 6 hidden layers with 120 neurons each. The available features for the source model include OWS feed flow rate and three DAF outlet properties (pH, COD, NH4-N). The activation function employed is hyperbolic tangent. The trends are visualized in the second row of Fig. 7, and the corresponding metrics are provided in the second column of Table 3. The test results, as shown in Fig. 7.d, indicate that they are not as satisfactory as those observed in the open-source model.

In the subsequent step, the first layer is constructed with the identity function and 5 neurons (total number of features and the output). The transferred layers are frozen, followed by an additional 3 layers with 30 neurons each. After completing the training procedure, fine-tuning is initiated with a learning rate of $10^{-5}$. The resulting model is termed "Industrial TL," and its trends are displayed in the second column of Fig. 8. Metrics for this model are provided in the last row of Table 1 and 2.

While the training performance is inferior to the standard model, both test and validation performances are enhanced. The Industrial TL model improves MSE by 20% for the test set and 47% for the validation set. Similarly, for MAE, the model enhances test and validation performance by 9% and 35%, respectively. The trends indicate notable improvements, particularly in the validation segment. The gap between the standard and actual values for the 725-800th days is closed when knowledge from the industrial source model is transferred to the target model. Consequently, the disparity between actual values and the standard model prediction diminishes around the 850$^{th}$ and 900$^{th}$ days. These results illustrate that despite the source model's smaller dataset size and potential noise in measurements, transfer learning effectively exploits the structural and locational similarities between the source and target models to enhance prediction performance. In many standard ANN or RNN studies, the validation step mismatches can be easily classified as measurement issues or unpredictable and unmeasured disturbances, but this study shows that many uncaught operation swings and trends changes can be captured well using transfer learning.

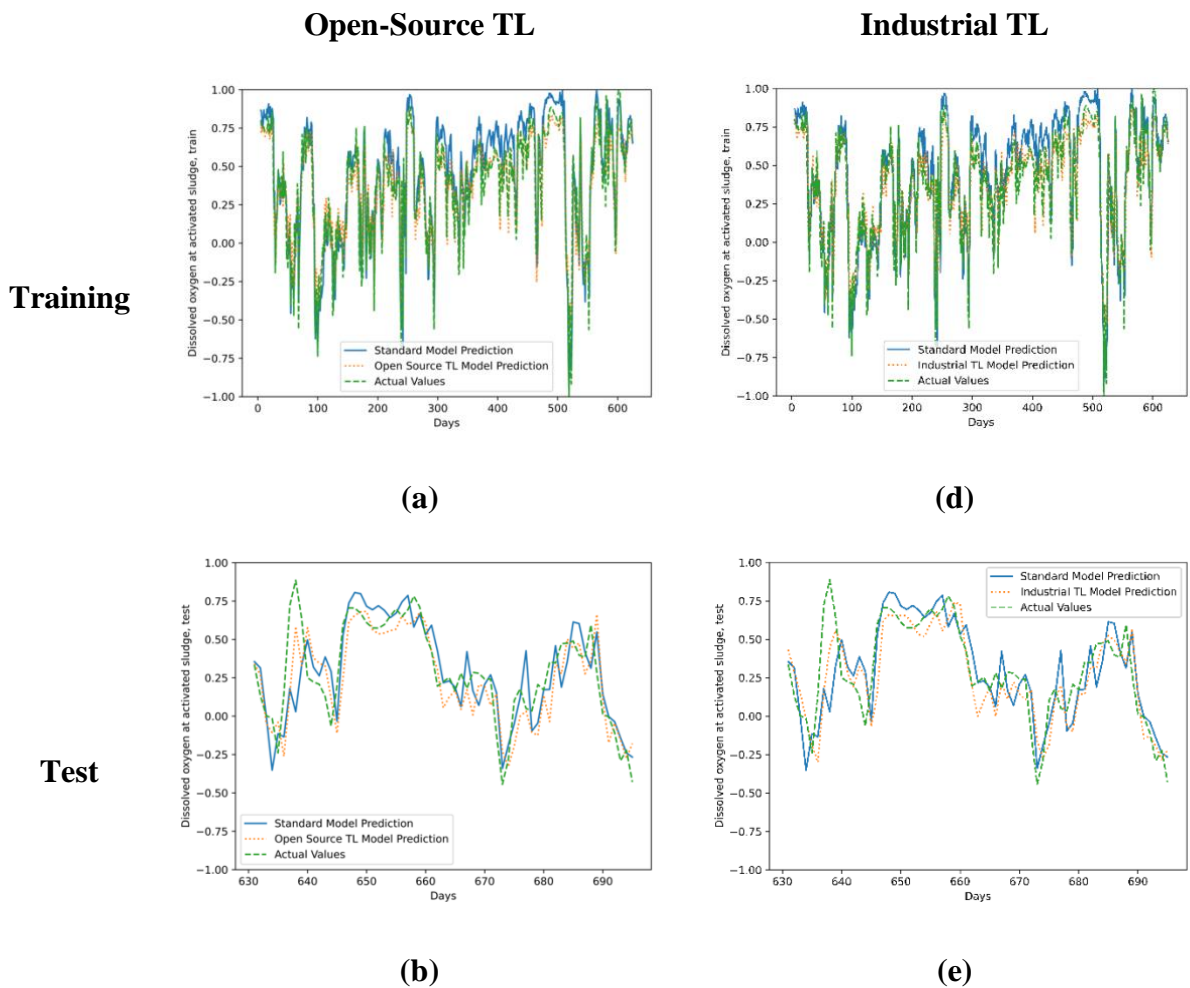

**Open-Source TL**          **Industrial TL**

Training — (a)     (d)

Test — (b)     (e)

**Validation** 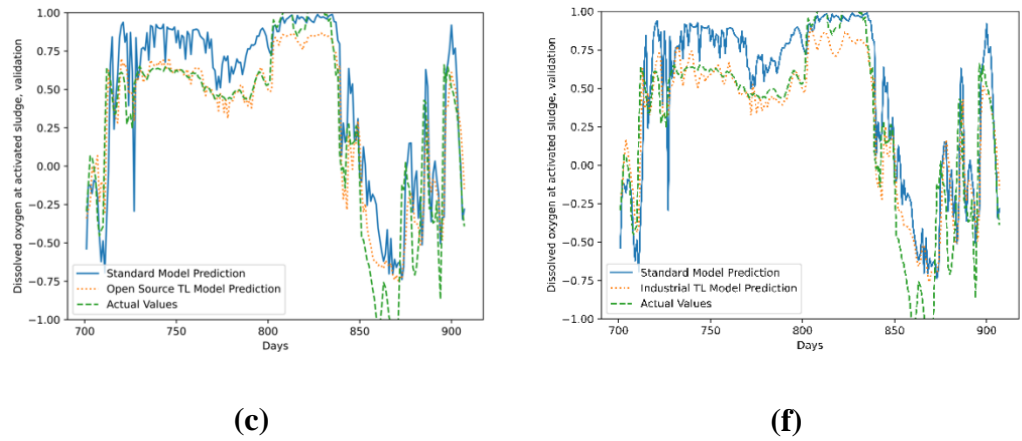

(c)            (f)

**Fig. 8.** Trends for open-source TL and industrial TL models.

**4.4. Physics-Informed Transfer Learning with an Industrial Source Plant**

The concluding case study integrates the transfer learning model from the preceding section with an industrial source plant and incorporates physics information derived from the open-source model. This enhanced model represents an upgraded version of the previous industrial transfer learning model, featuring a distinct objective function. Specifically, the physics loss associated with dissolved oxygen concentration in ASM1 [33] is integrated into the objective function, compelling the training process to adhere to the underlying physics principles. For detailed physics information, readers are referred to Eq. 1 and ASM1 [33].

In the physics-informed transfer learning model (PITL), the initial layer once again employs the identity activation function. Three layers are transferred from the industrial source model and held constant (frozen). Subsequently, three additional layers are appended to the model, resulting in a custom architecture that incorporates a physics-informed loss function. In contrast to previous transfer learning models, where custom layers were implemented as LSTM units due to their convergence to lower MSE, the PITL model yielded suboptimal results when LSTM layers were employed in this context. This outcome was anticipated, as the combination of transfer learning and a physics-informed loss function can increase model complexity. To address this, and to leverage the advantages of both transfer learning and physics-informed neural networks, the custom layers are structured as simple RNNs, which converge to lower MSE compared to LSTM and GRU structures.

The final PITL model exhibits lower test and validation MSE not only compared to the standard model but also in comparison to the industrial transfer learning model. The results presented in Table 1 indicate a 27% reduction in test MSE and a 59% reduction in validation MSE compared to the standard model. Similarly, MAE improvements for test and validation are 13% and 46%, respectively. Furthermore, in comparison to the industrial transfer learning model, the PITL model demonstrates improvements of 9% in test MSE, 23% in validation MSE, 4% in test MAE, and 17% in validation MAE.

The left column of Fig. 9 illustrates the trend comparison with the standard model, while the right column displays the comparison between the PITL and the industrial transfer learning model. Over the validation period, the PITL predictions closely align with actual values, particularly after the 800th day, showcasing the model's superior performance compared to both the standard model and the industrial transfer learning model.

These results highlight that the combination of a physics-informed objective function and transfer learning allows for a simplified model structure, utilizing simple RNN layers instead of more complex structures like LSTM. Consequently, this approach enhances the sequential prediction performance compared to classical machine learning and standard transfer learning. Importantly, despite the physics information and transferred model not being specific to the target model, the final model effectively leverages related information from both first principles and machine learning models. This promising outcome suggests that the integration of physics-informed neural networks and transfer learning, when applied to industrial data, presents a novel and effective solution to real-world challenges.

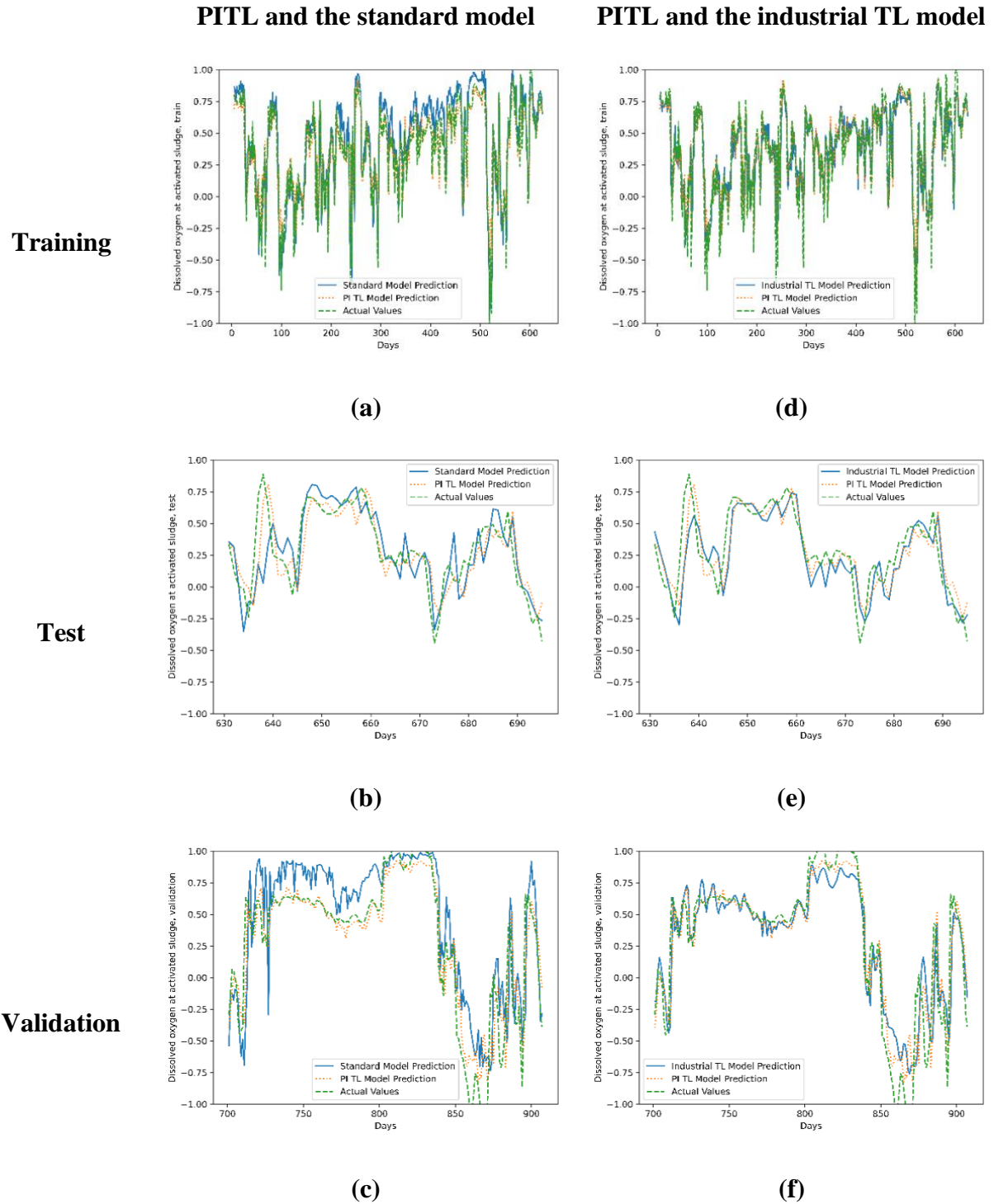

**Fig. 9.** Trends for PITL model.

## 5. Conclusion

In this work, utilization of transfer learning to regression problems related to complex and actual plants is discussed. The prediction performance of dissolved oxygen concentration at the activated sludge of an industrial refinery's wastewater treatment plant is increased by three transfer learning cases. The initial source model, ASM1, is an open-source simulation model, but its structure differs significantly from that of the target plant. The second source model represents another wastewater treatment plant within the same refinery. Despite having fewer available features and a smaller dataset, the model is compromised by noisy measurements due to real-life conditions. Despite these drawbacks in the two source models, the implementation of transfer learning has effectively boosted the prediction performance of the target plant.

Notably, a physics-informed transfer learning (PITL) model is devised, incorporating the layers of the industrial source model while introducing a customized objective function during the training process. This objective function integrates physics information derived from the open-source model. The resulting PITL model converges to the lowest MSE and MAE for both test and validation sets. Consequently, the validation MSE for the PITL model is approximately 60% lower than that of the classical machine learning model.

The study reveals the inadequacy of the classical dynamic (recurrent) machine learning approach, illustrating that choosing for either a less or more complex model resulted in underfitting and overfitting, respectively. Transfer learning, however, proves to be a viable solution, leveraging knowledge from related tasks, even in scenarios where the number of features prevents dataset integration. Furthermore, the combination of physics-based information derived from the open-source model and the machine learning model tailored to the target task yields the best model fit.

**Acknowledgements**

We gratefully acknowledge TÜPRAS¸ refinery and TÜPRAS¸ R&D department for their contributions and support.